\RequirePackage[T1]{fontenc}
\documentclass[letterpaper,10pt,conference]{ieeeconf}
\IEEEoverridecommandlockouts
\usepackage{amsmath,amssymb,graphicx,booktabs,array,multirow}
\usepackage{times,mathptmx}
\usepackage{capt-of}
\usepackage{algorithm,algpseudocode}
\usepackage{pifont}
\usepackage[table]{xcolor}
\usepackage{tikz}
\usetikzlibrary{arrows.meta,positioning,fit}
\usepackage[hidelinks]{hyperref}
\newcommand{\mt}{\widetilde m}
\newcommand{\ours}{RedirectMimic}
\newcommand{\ind}{\mathbf{1}}
\newcommand{\oracle}{RefEdit$^{\dagger}$}
\newcommand{\forget}{ForgetMimic$^{\ddagger}$}
\newcommand{\yes}{\ding{51}}
\newcommand{\no}{\ding{55}}
\title{RedirectMimic: Local Motion Content Redaction\\for Humanoid Whole-Body Control}
\author{Zepeng Wang and Xiaochuan Shi$^{\dagger}$\\[2pt]
{\normalsize Wuhan University}\\[2pt]
{\small $^{\dagger}$Corresponding author.}}
\IEEEaftertitletext{\begin{minipage}{\textwidth}
\centering
\includegraphics[width=\textwidth]{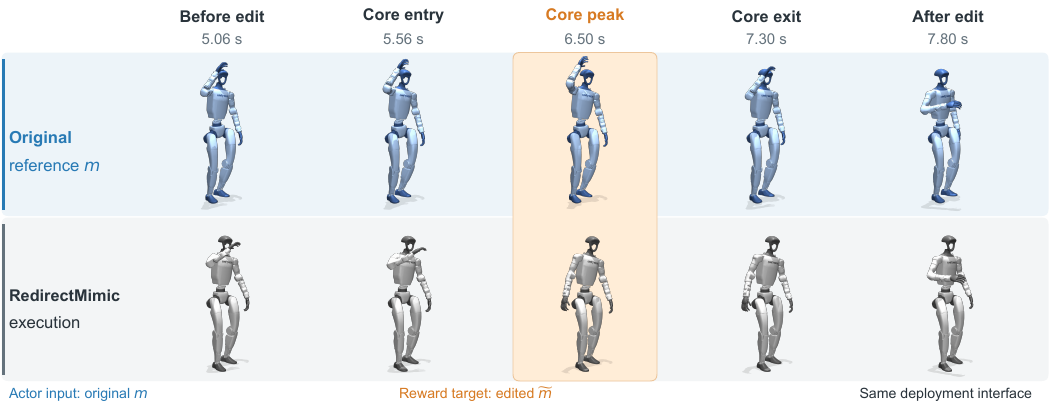}
\captionof{figure}{Local redaction in a selected G1 simulation rollout. Top: original reference $m$. Bottom: recorded \ours\ execution at matching times. The pre-edit column shows ordinary hand-to-mouth motion. At core entry, execution lags the edited target.}
\label{fig:teaser}
\end{minipage}
\vspace{10pt}}
\begin{document}
\bstctlcite{IEEEreferenceStyle}
\maketitle
\thispagestyle{plain}\pagestyle{plain}
\begin{abstract}
Humanoid motion interfaces should accommodate application-specific movement constraints while retaining useful reference behavior. Local content redaction requires a controller to replace selected movements even when it continues to receive the original reference. Removing the tracking incentive for those movements leaves their replacement unspecified. We introduce RedirectMimic, which separates the reference observed by the actor from the target used to train its execution. A rule-generated edited reference guides tracking rewards during proximal policy optimization, while the deployed observation and action interfaces remain unchanged. We evaluate pose- and speed-defined arm editing in Unitree G1 simulation using a tracker pretrained on 7,509 motion clips. On the overhead-arm task, RedirectMimic reduces core-window content occurrence to 2.7\%, compared with 58.9\% for reward zeroing and 32.8\% for anti-tracking. It achieves 94.2\% core-window joint success, combining content suppression, fall-free execution, and local retention. On the speed-defined task, fast outward arm movements occurred at similarly low rates under all three local editing methods. Across both tasks, RedirectMimic achieved the lowest replacement error among these methods and tracked unedited motions with accuracy comparable to that of the pretrained controller.
Project page: \href{https://xiaopeng-whu.github.io/RedirectMimic/}{\textcolor{magenta}{\nolinkurl{https://xiaopeng-whu.github.io/RedirectMimic/}}}.
\end{abstract}
\section{Introduction}
Motion tracking provides a practical interface between human demonstrations and humanoid control, including whole-body teleoperation and data collection~\cite{he2024omnih2o,ze2025twist2}. Physics-based imitation methods learn diverse motor skills from reference trajectories~\cite{peng2018deepmimic,liao2025beyondmimic,wang2026fast}, and large motion collections broaden the behaviors available to a controller~\cite{mahmood2019amass}. Faithful reproduction, however, is not always the desired deployment behavior. A reference may contain an arm movement that a particular application excludes, embedded within otherwise acceptable locomotion or gestures. This calls for \emph{selective} editing that replaces the specified movement while retaining the rest of the motion.

Figure~\ref{fig:teaser} illustrates this local objective. The controller must execute a suitable replacement while maintaining balance and the surrounding movement. Arm motion and balance are dynamically coupled: a local joint edit can affect torso and leg execution even when their reference coordinates remain unchanged. Useful redaction therefore combines content suppression, replacement fidelity, and physical retention.

Recent work studies selective behavior modification in humanoid policies~\cite{luan2026forgetmimic} and motion generators~\cite{dematteis2026hmu,wang2026safemo}, as well as learned policy redirection~\cite{wang2026uncloning}. Local humanoid editing couples this objective to the preservation of surrounding physical behavior. The resulting control question is \emph{how a tracker can learn a chosen replacement while its input continues to request the original movement}.

We address this question with \ours. An offline rule constructs a paired reference $(m,\mt)$, preserving the root trajectory and all unedited joint coordinates. During fine-tuning, the actor continues to observe $m$ and the robot state, while tracking rewards use $\mt$. Ordinary proximal policy optimization (PPO)~\cite{schulman2017ppo} updates the existing actor and critic. An edit-window start distribution increases exposure to the sparse replacement objective. At deployment, the actor retains the original observation and action interfaces.

The central distinction is between suppressing a movement and specifying its replacement. Reward zeroing removes the incentive to track the selected arm, whereas anti-tracking rewards deviation from it. Neither objective specifies where the arm should go. In G1 simulations of pose- and speed-defined edits, the arm followed the prescribed replacement more closely when that trajectory guided training.

Our contributions are threefold:
\begin{itemize}
\item \textbf{Local redaction under original-reference inputs.} We formulate selective motion replacement as policy editing, requiring the controller to retain surrounding motion while its input continues to specify the original behavior.
\item \textbf{Paired-reference policy fine-tuning.} We introduce \ours, combining rule-generated tracking targets and edit-window sampling to train local replacements under the original-reference observation interface.
\item \textbf{Evidence for explicit replacement targets.} Matched comparisons in G1 simulation show the highest overhead-arm joint success and lowest replacement error across both tasks among local methods. Trajectory analysis distinguishes faithful replacement from threshold suppression.
\end{itemize}

\section{Related Work}
\subsection{Humanoid Tracking and Physical Feasibility}
DeepMimic establishes reinforcement-learning imitation of reference skills~\cite{peng2018deepmimic}, while AMP learns motion-style rewards from demonstrations~\cite{peng2021amp}. Perpetual Humanoid Control extends imitation and recovery across large motion libraries~\cite{luo2023phc}. Recent robot systems pursue expressive upper-body imitation with locomotion in ExBody~\cite{cheng2024exbody}, and more general whole-body tracking in ExBody2~\cite{ji2024exbody2}. GMT combines adaptive sampling with motion-specialized experts~\cite{chen2025gmt}. SONIC studies scaling motion data, model capacity, and compute~\cite{luo2026sonic}. HOVER distills multiple control modes into one policy~\cite{he2024hover}, while TWIST combines reinforcement learning and behavior cloning for coordinated imitation~\cite{ze2025twist}. A recent empirical study examines design choices for general motion tracking~\cite{amadio2026tracking}. BeyondMimic provides the tracking formulation underlying our mjlab implementation~\cite{liao2025beyondmimic,zakka2026mjlab}. Our study modifies selected local responses of an existing tracker.

Reference fidelity can conflict with physical feasibility. ASAP adapts policies to simulation--reality discrepancies using a learned residual action model~\cite{he2025asap}. AMO combines trajectory optimization and reinforcement learning for adaptive control~\cite{li2025amo}, while OmniTrack generates physically consistent references before training a general tracker~\cite{li2026omnitrack}. PGMT relaxes tracking to accommodate terrain~\cite{li2026pgmt}. These approaches motivate validating a replacement dynamically. Here the reason to deviate is a fixed content rule, and the actor must continue observing the unedited reference.

\subsection{Motion Completion and Local Editing}
MoMask supports text-conditioned temporal inpainting through masked motion modeling~\cite{guo2024momask}. CondMDI supports sparse and partial keyframe constraints~\cite{cohan2024condmdi}, while DartControl generates motion autoregressively with text and spatial control~\cite{zhao2025dart}. MaskedMimic formulates physically simulated character control as motion inpainting from partial descriptions~\cite{tessler2024maskedmimic}. CLoSD closes the loop between a diffusion planner and a tracking controller~\cite{tevet2025closd}. Our input--target split brings local replacement into policy fine-tuning: the actor learns to execute the edited behavior from the original reference.

\subsection{Behavior Suppression and Policy Redirection}
ForgetMimic suppresses designated humanoid motions while retaining other motions~\cite{luan2026forgetmimic}. Local redaction additionally requires preservation within the targeted motion. Behavior Uncloning encodes redirection in policy weights without inference-time steering~\cite{wang2026uncloning}. For local humanoid editing, the desired response must specify both the affected body parts and their replacement trajectory.

Human Motion Unlearning suppresses unwanted generated motions through latent code replacement, including edits within composite behaviors~\cite{dematteis2026hmu}. SafeMo combines language-guided data refinement with continuous-space motion unlearning~\cite{wang2026safemo}. Our paired references specify a local replacement for closed-loop humanoid execution, where balance and surrounding motion must be preserved throughout the edit.

\section{Local Redaction with Paired References}
\begin{figure*}[!t]
\centering
\includegraphics[width=\textwidth]{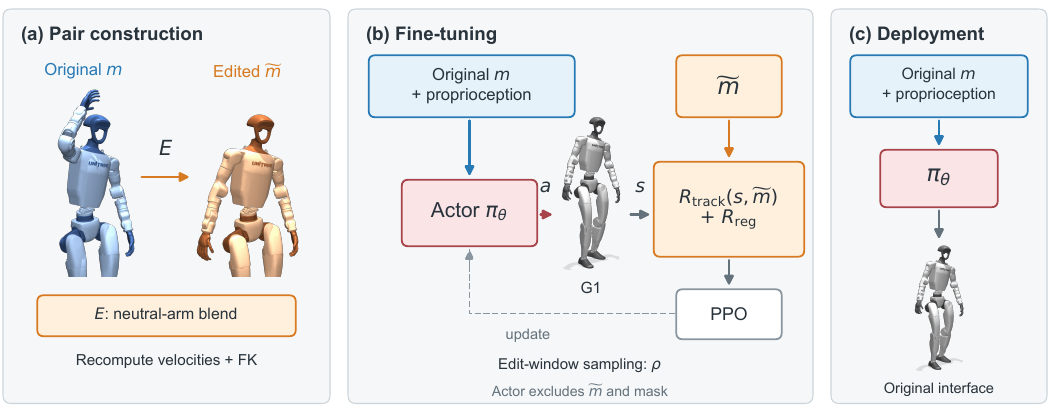}
\caption{\ours\ overview. (a) Construct the local replacement. (b) Fine-tune with original-reference observations and edited-target rewards. (c) Retain the deployment interface. G1 images show reference poses and recorded simulation execution.}
\label{fig:method}
\end{figure*}

Figure~\ref{fig:method} summarizes the three stages: offline pair construction, dual-reference fine-tuning, and deployment through the original policy interface.
\subsection{Problem and Information Constraint}
Let $m$ denote a reference motion, $m_t$ its frame at time $t$, and $\pi_0$ a pretrained motion tracker. An offline editor $E$ constructs a paired reference $\mt=E(m)$ by replacing a specified movement over a selected set of joints and frames. For clips that contain no such movement, $E(m)=m$. The control objective is to execute the replacement while preserving the surrounding motion and the ability to track unedited clips.

At time $t$, the actor observes the robot state $s_t$ and the original reference frame $m_t$,
\begin{equation}
o_t=\operatorname{Obs}(s_t,m_t),\qquad
a_t\sim\pi_\theta(\cdot\mid o_t).
\label{eq:obs}
\end{equation}
The deployed actor is a feedforward network with current-frame reference inputs and the previous action. It receives no future reference frames, clip identifier, edit mask, or edited-reference error. All reference-derived observations and control paths use $m$, and the content rule is fixed throughout training and deployment.

\subsection{Constructing Consistent Local Targets}
The editor uses a content rule to locate a movement in the original clip and a replacement target to specify what the selected joints should do there. The application determines which joints and motion features enter the rule. For each detected event, the editor defines a core interval $C$ for the replacement and an extended interval $I$ that includes the transitions into and out of the surrounding motion.

The motion feature follows from the constraint. A position threshold can identify entry into a restricted region. When the same path is acceptable at low speed, the rule can instead use velocity along the relevant direction to identify a rapid movement. Once the rule selects an interval, the application supplies the desired replacement for the affected joints.

Let $q_{t,j}$ be the original angle of joint $j$ and $q^\star_{t,j}$ its prescribed target. The editor blends the two on the original timeline,
\begin{equation}
\widetilde q_{t,j}=(1-w_{t,j})q_{t,j}+w_{t,j}q^\star_{t,j}.
\label{eq:blend}
\end{equation}
For a selected joint, the weight rises smoothly to one, stays at one throughout $C$, and returns to zero by the end of $I$. It is zero for all other joints and outside $I$. This preserves the root trajectory and every unedited joint coordinate while giving the tracker a continuous reference at the edit boundaries.

Changing a joint angle also moves its descendant bodies. We therefore recompute body poses by forward kinematics and derive joint and body velocities from the edited motion. The edit support $B$ includes the transitions and neighboring frames used to compute velocities. We check the resulting reference against the content rule and verify that the edit stays within its support. The resulting position, orientation, and velocity targets describe the same movement.

\subsection{Training Against the Edited Reference}
The fine-tuning reward is
\begin{equation}
r_t=\Delta t\left[
R_{\mathrm{track}}(s_t,\mt_t)+R_{\mathrm{reg}}(s_t,a_t)\right].
\label{eq:reward}
\end{equation}
Following the motion-tracking reward formulation of BeyondMimic~\cite{liao2025beyondmimic}, we evaluate anchor pose, relative body poses, and body linear and angular velocities against $\mt_t$. We retain the pretrained tracker's regularization terms and weights in $R_{\mathrm{reg}}$.

The edited reference supplies one full-body tracking target. During the edit, the selected bodies follow the replacement and the unaffected bodies follow their original targets. Outside the edit support and throughout unedited clips, the reward trains the policy to track the original motion.

The policy optimizes the full-body action jointly. A local change in the reference can require coordinated adjustments elsewhere in the body to maintain balance. Reference locality specifies the desired motion, and closed-loop optimization learns how to execute it with physical regularization. PPO updates both actor and critic under this objective.

Both references advance on a shared timeline and use the same anchor alignment. The reward therefore evaluates the replacement at the phase represented by the actor's original input. Episodes initialize from $m$, so training includes reaching the replacement from the original motion state.

\subsection{Sampling Sparse Edit Intervals}
Adaptive motion sampling is established in general tracking~\cite{chen2025gmt}. Here, failure-driven sampling need not emphasize intervals where following $m$ remains physically feasible but produces the wrong content. We therefore mix the base start distribution $Q_{\mathrm{base}}$ with uniform sampling over an edit-start pool $\mathcal P$. The pool contains frames where the selected joints differ between the two references and a preceding lead-in within the same clip. If its fraction of the motion timeline is $p_{\mathrm{pool}}$, the start distribution is
\begin{equation}
\begin{aligned}
Q&=(1-\rho)Q_{\mathrm{base}}+\rho\mathcal U(\mathcal P),\\
\rho&=\min(\rho_{\max},\kappa p_{\mathrm{pool}}).
\end{aligned}
\label{eq:sampling}
\end{equation}
The factor $\kappa$ increases exposure to sparse replacements, while the cap $\rho_{\max}$ reserves training coverage for the rest of the motion library. The lead-in exposes the policy to the transition from the original movement into the edited interval. With an empty edit pool, sampling reduces to $Q_{\mathrm{base}}$.

The base sampler combines failure-rate weighting with uniform coverage. Its feedback uses base-sampled episodes exclusively, keeping its estimate of tracking difficulty independent of the extra edit-window starts. This separation controls exposure to the editing objective while preserving adaptation to difficult motions across the library.

Algorithm~\ref{alg:redirectmimic} gives the pseudocode for \ours.

\begin{algorithm}[!ht]
\caption{RedirectMimic}
\label{alg:redirectmimic}
\small
\begin{algorithmic}[1]
\Require Motions $\mathcal M$, editor $E$, pretrained actor and critic $(\theta_0,\phi_0)$, sampler $Q_{\mathrm{base}}$, update/rollout lengths $K,H$
\Ensure Edited policy $\pi_\theta$
\State Initialize $(\theta,\phi)\gets(\theta_0,\phi_0)$
\State Pair targets $\widetilde{\mathcal M}\gets\{m\mapsto E(m):m\in\mathcal M\}$
\State Build edit-start pool $\mathcal P$ and set $\rho$ (Eq.~\eqref{eq:sampling})
\For{PPO update $k=1,\ldots,K$}
  \State Clear rollout buffer $\mathcal D\gets\varnothing$
  \For{rollout step $h=1,\ldots,H$}
    \If{first step or previous episode ended}
      \State Mix starts $Q\gets\rho\mathcal U(\mathcal P)+(1-\rho)Q_{\mathrm{base}}$
      \State Sample $(m,t)\sim Q$, retrieve $\mt\gets\widetilde{\mathcal M}[m]$
      \State $s_t\gets\operatorname{Init}(m_t)$ \Comment{Start from the original motion}
    \EndIf
    \State Observe original $o_t\gets\operatorname{Obs}(s_t,m_t)$
    \State Act $a_t\sim\pi_\theta(\cdot\mid o_t)$
    \State Step simulator with $a_t$ to obtain $s_{t+1}$ and done flag $d_t$
    \State Compute $r_t$ by Eq.~\eqref{eq:reward} \Comment{Track $\mt_t$}
    \State Collect $(o_t,a_t,r_t,s_{t+1},d_t)$ in $\mathcal D$
    \State Advance reference frame $t\gets t+1$
  \EndFor
  \State Estimate advantages and returns $(\widehat A,\widehat V)$ with GAE on $\mathcal D$ using critic $V_\phi$
  \State Update $(\theta,\phi)\gets\operatorname{PPO}(\theta,\phi,\mathcal D,\widehat A,\widehat V)$
\EndFor
\State \Return $\pi_\theta(\cdot\mid\operatorname{Obs}(s,m))$
\end{algorithmic}
\end{algorithm}

\section{Experiments}
In the experimental section, we aim to answer the following key research questions.

\noindent\textbf{Q1 (Editing effectiveness).} Can the controller suppress a specified movement while staying upright and preserving the surrounding motion?

\noindent\textbf{Q2 (Motion retention).} How closely does the fine-tuned controller track motions that require no edit?

\noindent\textbf{Q3 (Replacement fidelity).} How closely does the edited limb follow the prescribed trajectory when different controllers suppress the same movement?

\noindent\textbf{Q4 (Training choices).} How do edit-window sampling and initialization from a pretrained tracker affect learning the replacement?

\subsection{Experiment Setup}
We consider two arm-editing scenarios for the 29-joint Unitree G1 in mjlab. S1 replaces rapid outward extensions when wrist speed along the shoulder-to-wrist axis exceeds $1.5\,\mathrm{m/s}$. S2 replaces overhead reaches when the wrist rises more than $0.2\,\mathrm{m}$ above the shoulder. In both scenarios, the edited reference brings the selected arm to a neutral pose with the upper arm near the torso and the elbow bent. Both training sets draw from the same 7,509 clips in Retargeted AMASS for Robotics~\cite{zhaoamass,mahmood2019amass}. S1 uses the full pool. S2 retains 6,650 clips after excluding borderline wrist heights and edited references that still contain an overhead reach.

We evaluate all 173 eligible S1 windows and 149 S2 windows sampled across motion sources and screened for reference feasibility. Each window includes 16 randomized rollouts per controller with matched starting states before the edit.

We measure content occurrence, replacement error, preserved-body error, and joint success. For each rollout, $D_C$ denotes the event that the selected arm exceeds the task threshold for a cumulative duration of at least $0.06\,\mathrm{s}$ within core interval $C$. The content occurrence rate (Content) is the fraction of all rollouts in which $D_C$ occurs, including rollouts with falls in the evaluation pool. A fall $F_H$ occurs when pelvis height drops below $0.35\,\mathrm{m}$ anywhere in horizon $H$. With preserved-body error $e_C^{\mathrm{keep}}$ measured relative to $m$ in the torso yaw frame, joint success is
\begin{equation}
J_C=\ind[\neg D_C]\cdot\ind[\neg F_H]
\cdot\ind[e_C^{\mathrm{keep}}\leq0.10\,\mathrm{m}].
\label{eq:joint}
\end{equation}
Arm error to $\mt$ measures replacement fidelity. Preserved-body errors within and outside the edit assess local retention.

\subsection{Baselines}
\label{sec:baselines}
Table~\ref{tab:properties} compares the editing mechanisms and deployment inputs of all evaluated methods.

\begin{table}[!ht]
\centering\footnotesize
\caption{Editing mechanisms and deployment inputs.}
\label{tab:properties}
\setlength{\tabcolsep}{4pt}
\begin{tabular}{lcccc}\toprule
Method & \shortstack{Local\\edit} & \shortstack{Explicit\\target} & \shortstack{Policy\\edit} & \shortstack{Original\\input}\\\midrule
Pretrained & \no & \no & \no & \yes\\
Vanilla-FT & \no & \no & \no & \yes\\
\forget & \no & \no & \yes & \yes\\
AntiTrack & \yes & \no & \yes & \yes\\
ZeroTrack & \yes & \no & \yes & \yes\\
\rowcolor{black!10}\textbf{\ours} & \yes & \yes & \yes & \yes\\\midrule
\oracle & \yes & \yes & \no & \no\\\bottomrule
\end{tabular}
\par\smallskip
\parbox{\columnwidth}{Local edit: a body- and time-specific intervention. Explicit target: a specified replacement trajectory. Policy edit: the editing objective is learned in policy weights. Original input: deployment uses $m$.}
\end{table}

\textbf{Pretrained and Vanilla-FT.} Pretrained evaluates the original tracker $\pi_0$. Vanilla-FT continues training on the original tracking objective to control for additional optimization. We evaluate the same S1-trained Vanilla-FT policy on both tasks.

\textbf{\forget.} We adapt ForgetMimic's clip-level anti-reward objective~\cite{luan2026forgetmimic} to the shared tracker and training protocol. Every frame of a clip containing edited content receives an added $-\lambda\bar r_t$, where $\bar r_t$ is the weight-normalized tracking reward toward $m$. Setting $\lambda$ to the total tracking weight cancels tracking reward on these clips. Other clips keep the original objective, and $\mt$ is not used. This reimplementation uses the shared sampler, regularizers, and termination settings below. The original method uses uniform motion/frame sampling and an additional survival bonus.

\textbf{ZeroTrack and AntiTrack.} These local reward controls isolate reward suppression from explicit replacement. Both operate on four body-level tracking terms. For a term with squared error $e_{t,b}$, scale $\sigma$, and $N=14$ tracked bodies, the original reward is $\exp[-\sum_b e_{t,b}/(N\sigma^2)]$. Let $A_t$ contain tracked descendants of the edited arm joints. Inside the edit support, ZeroTrack uses
\begin{equation}
r_t^{\mathrm{zero}}=
\exp\!\left[-\frac{\sum_{b\notin A_t}e_{t,b}}{N\sigma^2}\right].
\label{eq:zero}
\end{equation}
The original denominator $N$ is retained, so ``zeroing'' removes selected error contributions from the coupled exponential reward. AntiTrack adds
\begin{equation}
r_t^{\mathrm{anti}}=r_t^{\mathrm{zero}}-
\exp\!\left[-\frac{\sum_{b\in A_t}e_{t,b}}{|A_t|\sigma^2}\right].
\end{equation}
Both use $m$ for these errors and equal the original tracking terms outside the edit support. Anchor rewards and physical regularizers remain unchanged. AntiTrack uses unit anti-reward weight.

\textbf{\oracle\ (oracle).} This reference supplies $\mt$ directly to $\pi_0$ after initialization from the common original-reference state. It assesses target feasibility with deployment-time access to the edited reference, under a different input condition from Eq.~\eqref{eq:obs}.

\forget, AntiTrack, ZeroTrack, and \ours\ share the pretrained checkpoint, task-specific data, architecture, start distribution, and 15,000-iteration update budget. Observation normalizers update across all fine-tuned methods.

\begin{figure*}[!t]
\centering
\includegraphics[width=\textwidth]{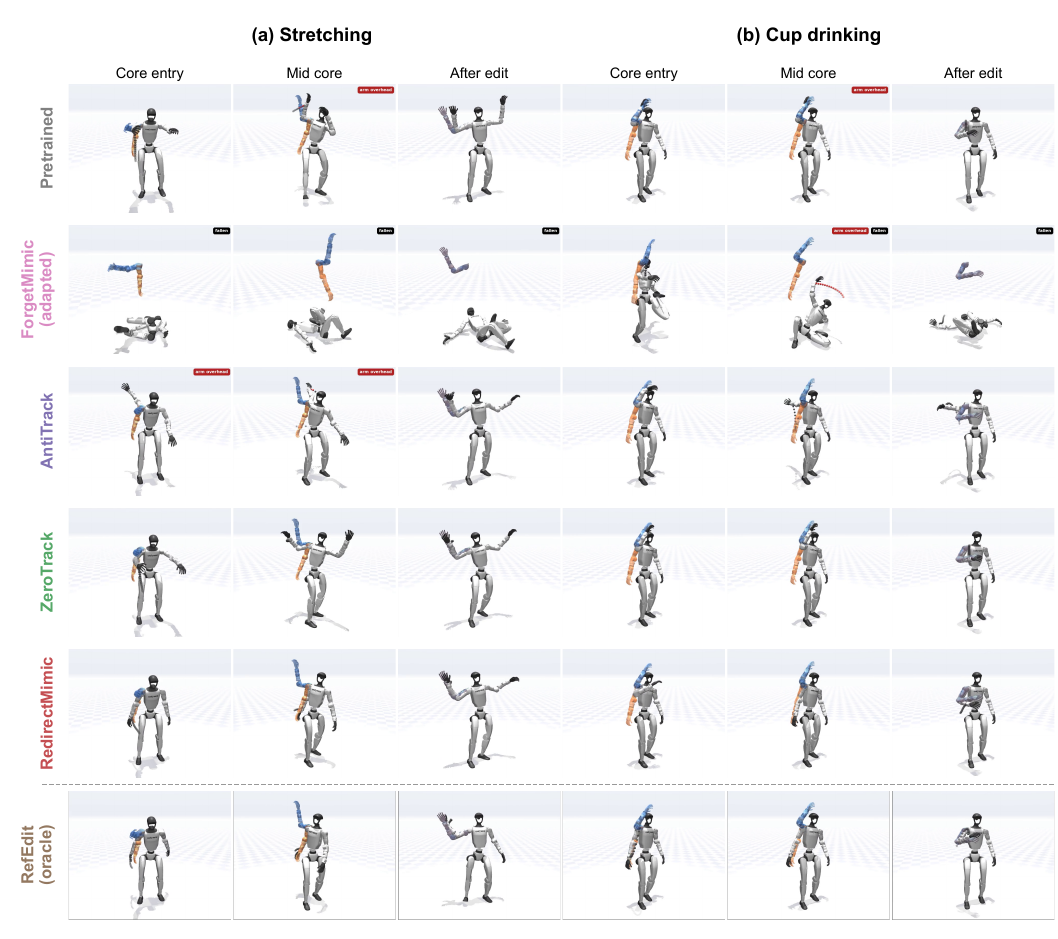}
\caption{Stretching (a) and cup drinking (b) at core entry, mid-core, and after the edit. The blue arm shows the original reference $m$ and the orange arm shows the edited target $\mt$.}
\label{fig:frames}
\end{figure*}

\subsection{Local Editing Performance}
To answer \textbf{Q1}, we compare content occurrence and joint success across controllers. As shown in Table~\ref{tab:main}, \ours\ achieved the lowest overhead-content occurrence and the highest joint success among the matched local methods. ZeroTrack removes the selected arm's tracking objective, leaving other reward terms to favor motion close to the original reference. AntiTrack rewards departure from that reference. Tracking $\mt$ additionally directs the arm toward the chosen replacement.

\begin{table}[!ht]
\centering\footnotesize
\caption{Local replacement performance. Rollout-mean rates and target errors (m).}
\label{tab:main}
\setlength{\tabcolsep}{3pt}
\begin{tabular}{lrrrr}\toprule
Method & $J_C$ $\uparrow$ & Content $\downarrow$ & Error $\downarrow$ & Fall $\downarrow$\\\midrule
\multicolumn{5}{l}{\textit{S2: overhead-arm replacement}}\\
Pretrained & 0.245 & 0.737 & 0.238 & 0.014\\
Vanilla-FT & 0.218 & 0.774 & 0.237 & \underline{0.011}\\
\forget & $<0.001$ & \underline{0.258} & 0.412 & 1.000\\
AntiTrack & \underline{0.650} & 0.328 & \underline{0.171} & \textbf{0.009}\\
ZeroTrack & 0.408 & 0.589 & 0.224 & \underline{0.011}\\
\rowcolor{black!10}\textbf{\ours} & \textbf{0.942} & \textbf{0.027} & \textbf{0.079} & 0.022\\
\cmidrule(lr){1-5}
\oracle & 0.972 & 0.003 & 0.046 & 0.013\\
\midrule
\multicolumn{5}{l}{\textit{S1: speed-defined replacement}}\\
Pretrained & 0.725 & 0.105 & 0.177 & 0.089\\
Vanilla-FT & 0.745 & 0.120 & 0.177 & 0.079\\
\forget & 0.000 & 0.031 & 0.416 & 1.000\\
AntiTrack & 0.857 & \underline{0.023} & 0.174 & \underline{0.061}\\
ZeroTrack & \textbf{0.897} & \textbf{0.020} & \underline{0.158} & \textbf{0.049}\\
\rowcolor{black!10}\textbf{\ours} & \underline{0.874} & \underline{0.023} & \textbf{0.143} & 0.064\\
\cmidrule(lr){1-5}
\oracle & 0.803 & 0.000 & 0.079 & 0.103\\
\midrule
\bottomrule\end{tabular}
\par\smallskip
\parbox{\columnwidth}{$^{\dagger}$Oracle receives $\mt$. $^{\ddagger}$ForgetMimic objective adapted to the shared protocol~\cite{luan2026forgetmimic}.}
\end{table}

The stretching example in Fig.~\ref{fig:frames}(a) makes this distinction visible. The local reward controls retain upward or lateral arm responses, while \ours\ lowers the edited arm toward the replacement during the core. In the cup-drinking example in Fig.~\ref{fig:frames}(b), ZeroTrack satisfies the content threshold with the hand near its original pose. \ours\ brings the hand closer to the lowered target. The entry and post-edit frames show the surrounding transitions.

The lower aggregate replacement error confirms closer agreement with the intended alternative. The S2 joint-success advantage persists despite a higher fall rate than AntiTrack and remains under unfiltered evaluation.

Under the shared-protocol adaptation, \forget\ yielded near-zero joint success, with falls dominating both tasks. Although its S2 content occurrence was lower than that of either local reward control, it rarely stayed upright through the edit. Joint success counts suppression only when the robot also stays upright and tracks the unedited body parts within tolerance.

S1 separates replacement fidelity from threshold suppression. The local methods suppressed fast outward movement similarly, with ZeroTrack attaining the highest joint-success point estimate. ZeroTrack's arm trajectory remained close to the original reference, while \ours\ more closely followed $\mt$. Slowing a movement can satisfy the speed rule along the original spatial path. Explicit targets therefore improve the requested replacement even when binary suppression is similar.

The \oracle\ reference closely tracked $\mt$ and nearly eliminated content in both tasks, supporting the feasibility of the replacement targets. \ours\ approached its overhead-arm joint performance while observing the original reference. In S1, accurate oracle arm tracking coexisted with more falls, showing why target fidelity and physical execution must be assessed together.

\begin{figure*}[!t]
\centering
\includegraphics[width=\textwidth]{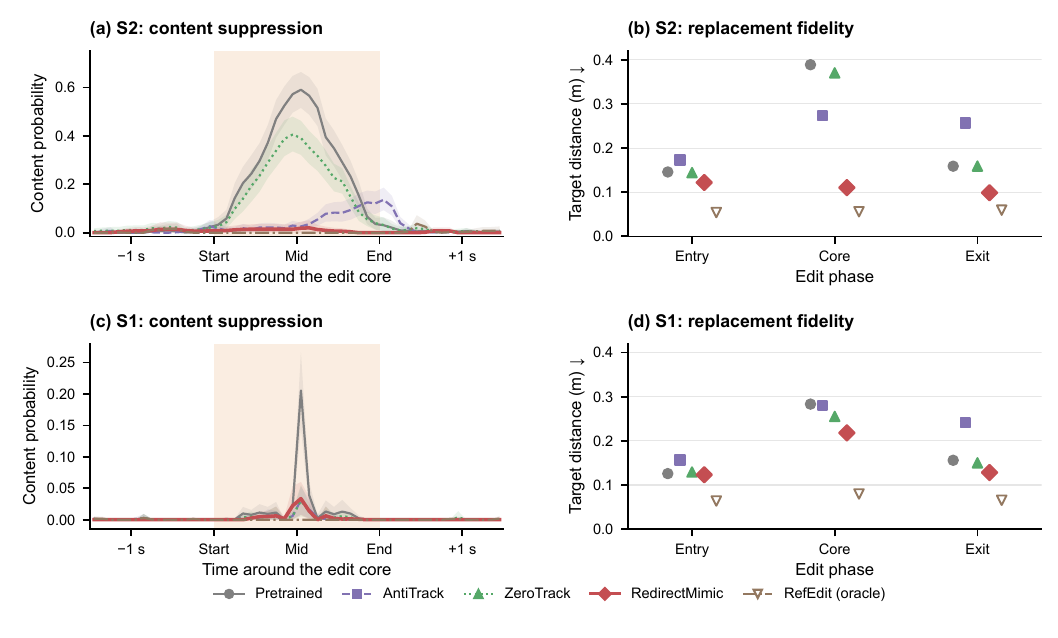}
\caption{Content suppression and replacement fidelity. The top and bottom rows show S2 and S1, respectively. The left panels show per-frame content probability with phase-normalized core time and clip-bootstrap intervals. The right panels show mean elbow--wrist distance to $\mt$ during entry, core, and exit. All panels use nonfallen rollouts. \forget\ is omitted because none remain in S1 and one remains in S2. S2 uses 149 windows in both columns. S1 uses 164 feasibility-filtered windows for the curves and all 173 primary windows for target distance.}
\label{fig:curves}
\end{figure*}

\subsection{Normal-Motion Retention}
To answer \textbf{Q2}, we evaluate each controller on 200 unedited clips per task. Table~\ref{tab:normal} shows that \ours\ tracked these motions with errors and fall rates close to those of the pretrained tracker. These clips retain their original tracking targets during fine-tuning.

\begin{table}[!ht]
\centering\footnotesize
\caption{Normal-motion retention. Error is retained-body distance to $m$ (m).}
\label{tab:normal}
\setlength{\tabcolsep}{4pt}
\begin{tabular}{lrrrr}\toprule
 & \multicolumn{2}{c}{S1} & \multicolumn{2}{c}{S2}\\
Method & Fall $\downarrow$ & Error $\downarrow$ & Fall $\downarrow$ & Error $\downarrow$\\\midrule
Pretrained & 0.098 & 0.070 & \underline{0.095} & \underline{0.070}\\
Vanilla-FT & \textbf{0.090} & \textbf{0.067} & 0.098 & \underline{0.070}\\
\forget & 0.131 & 0.078 & 0.366 & 0.151\\
AntiTrack & \textbf{0.090} & \textbf{0.067} & 0.103 & 0.073\\
ZeroTrack & \underline{0.093} & \underline{0.068} & \textbf{0.093} & \textbf{0.067}\\
\rowcolor{black!10}\textbf{\ours} & \underline{0.093} & \textbf{0.067} & 0.098 & \underline{0.070}\\
\bottomrule\end{tabular}\end{table}

Vanilla-FT improved aspects of normal tracking while continuing to reproduce the selected content. The matched local comparisons isolate the benefit of the replacement objective under a common optimization budget. The adapted whole-motion baseline increased normal-motion falls, particularly in S2, while the local objectives retained tracking close to pretrained behavior.

\subsection{Trajectory Redirection}
To answer \textbf{Q3}, we compare the distance to the target with the fraction of the intended displacement achieved along the edit direction. For executed elbow and wrist positions $x$, this fraction is
\begin{equation}
\alpha=\frac{\sum_{t,b}\langle x_{t,b}-m_{t,b},\mt_{t,b}-m_{t,b}\rangle}
{\sum_{t,b}\|\mt_{t,b}-m_{t,b}\|^2}.
\label{eq:alpha}
\end{equation}
Positions use the torso yaw frame, and sums pool each phase across nonfallen rollouts. The value $\alpha=1$ denotes the intended displacement along the edit direction. All values in Table~\ref{tab:alpha} are below one, so larger values indicate closer agreement along this direction. Target distance also captures error perpendicular to the edit direction.

Figure~\ref{fig:curves} pairs content suppression with replacement fidelity. In S2, \ours\ keeps content probability low through the core and has the lowest target distance among the matched local methods in each edit phase. The largest separation occurs in the core, where the intended replacement differs most from the original movement. Table~\ref{tab:alpha} confirms that the executed displacement follows the edit direction. ZeroTrack shows little directed redirection, while AntiTrack departs from the original motion with a larger target error.

\begin{table}[!ht]
\centering\footnotesize
\caption{Core redirection on nonfallen rollouts. Error is elbow--wrist distance to $\mt$ (m). For \forget, -- indicates no nonfallen rollouts in S1 and one in S2.}
\label{tab:alpha}
\setlength{\tabcolsep}{4pt}
\begin{tabular}{lrrrr}\toprule
 & \multicolumn{2}{c}{S1} & \multicolumn{2}{c}{S2}\\
Method & $\alpha$ $\uparrow$ & Error $\downarrow$ & $\alpha$ $\uparrow$ & Error $\downarrow$\\\midrule
Pretrained & 0.12 & 0.283 & 0.05 & 0.389\\
\forget & -- & -- & -- & --\\
AntiTrack & \textbf{0.64} & 0.280 & \underline{0.71} & \underline{0.273}\\
ZeroTrack & 0.20 & \underline{0.255} & 0.12 & 0.370\\
\rowcolor{black!10}\textbf{\ours} & \underline{0.35} & \textbf{0.218} & \textbf{0.84} & \textbf{0.110}\\
\midrule
\oracle & 0.92 & 0.080 & 0.97 & 0.056\\
\bottomrule\end{tabular}\end{table}

S1 makes the complementary point. The local methods have similarly low content probability, while \ours\ stays closer to the edited trajectory through the core and exit transition. Table~\ref{tab:alpha} shows that AntiTrack moves farther along the edit direction in S1 yet has a larger mean target distance over the core than \ours. Jointly assessing suppression and target distance identifies whether a controller executes the requested alternative.

\subsection{Ablation Studies}
To answer \textbf{Q4}, we first isolate the effect of edit-window sampling on one LAFAN1 arm-mirroring motion. Without targeted sampling, both scratch training and fine-tuning retained the original content. Adding edit-pool starts enabled both to learn the replacement. Repeated exposure to the edited interval was therefore necessary in this single-pair experiment.

The S1 initialization comparison tests reuse of an existing tracker. Both scratch training and fine-tuning learned local replacements, with fine-tuning yielding closer target agreement and similar joint success relative to clip-level uncertainty. This supports fine-tuning as a route for modifying an available repertoire. The comparison treats the pretrained tracker as an existing resource, so total training costs differ.

\section{Discussion}
\label{sec:limits}
\textbf{An explicit replacement target guides the arm even when several trajectories satisfy the same content rule.} On the speed-defined task, slowing the wrist along its original path can satisfy the rule. The local methods achieved similar suppression rates, but \ours\ tracked the prescribed trajectory more closely. On the overhead-arm task, this gain in accuracy coincided with higher joint success.

\textbf{Local editing also requires the robot to maintain balance and preserve the surrounding movement.} Under the shared training protocol, the adapted ForgetMimic objective reduced content occurrence on edited clips, yet the robot often fell before the edit began. In \ours, the paired references preserve tracking targets for unaffected body parts. On unedited clips, tracking accuracy remained close to that of the pretrained controller.

\textbf{The original reference interface allows the policy to execute a learned replacement, but its timing matters for streaming control.} The offline editor uses the full clip to place transitions, while the actor must initiate them from current-frame inputs. The results on known clips in simulation motivate measuring transition timing over a full streaming rollout, including whether the selected movement occurs before or after the intended edit interval.

\section{Conclusion}
We present \ours, an approach to local humanoid motion redaction that combines paired-reference fine-tuning with edit-window sampling. Training against an edited reference while observing the original motion lets the policy learn selected replacements through its existing control interface. G1 simulations demonstrate pose- and speed-defined arm editing. On unedited clips, tracking errors and fall rates remained close to those of the pretrained tracker. Among the matched local editing methods, \ours\ achieved the highest joint success on the overhead-arm task and the lowest replacement error on both tasks. Trajectory analysis distinguishes suppression of the specified content from faithful execution of the replacement, supporting the use of explicit targets. Future work will extend evaluation to streaming references and physical humanoids, with emphasis on transition timing and whole-body stability.

\bibliographystyle{IEEEtran}
\bibliography{references}

@article{peng2018deepmimic,
  author={Xue Bin Peng and Pieter Abbeel and Sergey Levine and Michiel van de Panne},
  title={{DeepMimic}: Example-Guided Deep Reinforcement Learning of Physics-Based Character Skills},
  journal={ACM Transactions on Graphics},
  volume={37},
  number={4},
  pages={143:1--143:14},
  year={2018},
  doi={10.1145/3197517.3201311},
  url={https://doi.org/10.1145/3197517.3201311}
}

@article{liao2025beyondmimic,
  author={Qiayuan Liao and Takara E. Truong and Xiaoyu Huang and Yuman Gao and Guy Tevet and Koushil Sreenath and C. Karen Liu},
  title={{BeyondMimic}: From Motion Tracking to Versatile Humanoid Control via Guided Diffusion},
  journal={Science Robotics},
  volume={11},
  number={117},
  year={2026},
  doi={10.1126/scirobotics.adx8924},
  note={Art. no. eadx8924},
  url={https://doi.org/10.1126/scirobotics.adx8924}
}

@inproceedings{mahmood2019amass,
  author={Naureen Mahmood and Nima Ghorbani and Nikolaus F. Troje and Gerard Pons-Moll and Michael J. Black},
  title={{AMASS}: Archive of Motion Capture as Surface Shapes},
  booktitle={Proceedings of the IEEE/CVF International Conference on Computer Vision},
  pages={5441--5450},
  year={2019},
  doi={10.1109/ICCV.2019.00554},
  url={https://doi.org/10.1109/ICCV.2019.00554}
}

@misc{luan2026forgetmimic,
  author={Xukun Luan and Zhongxiang Lei and Chen Gong and Shaowei Li and Yuanguo Bi and Jinyan Liu},
  title={{ForgetMimic}: Motion Unlearning for Reinforcement Learning Humanoid Control},
  year={2026},
  howpublished={arXiv:2609.28378},
  url={https://arxiv.org/abs/2609.28378}
}

@inproceedings{dematteis2026hmu,
  author={De Matteis, Edoardo and Matteo Migliarini and Alessio Sampieri and Indro Spinelli and Fabio Galasso},
  title={Human Motion Unlearning},
  booktitle={Proceedings of the AAAI Conference on Artificial Intelligence},
  volume={40},
  number={5},
  pages={3533--3541},
  year={2026},
  doi={10.1609/aaai.v40i5.37351},
  url={https://ojs.aaai.org/index.php/AAAI/article/view/37351}
}

@inproceedings{wang2026safemo,
  author={Yiling Wang and Zeyu Zhang and Yiran Wang and Hao Tang},
  title={{SafeMo}: Linguistically Grounded Unlearning for Trustworthy Text-to-Motion Generation},
  booktitle={Findings of the Association for Computational Linguistics: EMNLP 2026},
  pages={2562--2585},
  year={2026},
  publisher={Association for Computational Linguistics},
  doi={10.18653/2026.findings-emnlp.126},
  url={https://aclanthology.org/2026.findings-emnlp.126/}
}

@inproceedings{wang2026uncloning,
  author={Hao Wang and Jiuzhou Lei and Dayou Li and Bangya Liu and Minghui Zheng and Manling Li and Ruohan Zhang and Zhiwen Fan},
  title={Behavior Uncloning: Distilling Mode Redirection into Policy Weights without Inference-Time Steering},
  booktitle={Conference on Robot Learning (CoRL)},
  year={2026},
  note={Accepted for publication},
  url={https://openreview.net/forum?id=Tjg1ZvCdXU}
}

@misc{schulman2017ppo,
  author={John Schulman and Filip Wolski and Prafulla Dhariwal and Alec Radford and Oleg Klimov},
  title={Proximal Policy Optimization Algorithms},
  year={2017},
  howpublished={arXiv:1707.06347},
  url={https://arxiv.org/abs/1707.06347}
}

@article{peng2021amp,
  author={Xue Bin Peng and Ze Ma and Pieter Abbeel and Sergey Levine and Angjoo Kanazawa},
  title={{AMP}: Adversarial Motion Priors for Stylized Physics-Based Character Control},
  journal={ACM Transactions on Graphics},
  volume={40},
  number={4},
  pages={144:1--144:20},
  year={2021},
  doi={10.1145/3450626.3459670},
  url={https://doi.org/10.1145/3450626.3459670}
}

@inproceedings{luo2023phc,
  author={Zhengyi Luo and Jinkun Cao and Alexander Winkler and Kris Kitani and Weipeng Xu},
  title={Perpetual Humanoid Control for Real-time Simulated Avatars},
  booktitle={Proceedings of the IEEE/CVF International Conference on Computer Vision},
  pages={10861--10870},
  year={2023},
  doi={10.1109/ICCV51070.2023.01000},
  url={https://doi.org/10.1109/ICCV51070.2023.01000}
}

@misc{zakka2026mjlab,
  author={Kevin Zakka and Qiayuan Liao and Brent Yi and Le Lay, Louis and Koushil Sreenath and Pieter Abbeel},
  title={{mjlab}: A Lightweight Framework for {GPU}-Accelerated Robot Learning},
  year={2026},
  howpublished={arXiv:2601.22074},
  url={https://arxiv.org/abs/2601.22074}
}

@misc{zhaoamass,
  author={Kun Zhao},
  title={Retargeted {AMASS} for Robotics},
  howpublished={Hugging Face dataset},
  note={Accessed: Oct. 26, 2026},
  url={https://huggingface.co/datasets/fleaven/Retargeted_AMASS_for_robotics}
}

@inproceedings{cheng2024exbody,
  author={Cheng, Xuxin and Ji, Yandong and Chen, Junming and Yang, Ruihan and Yang, Ge and Wang, Xiaolong},
  title={Expressive Whole-Body Control for Humanoid Robots},
  booktitle={Proceedings of Robotics: Science and Systems},
  year={2024},
  doi={10.15607/RSS.2024.XX.107},
  url={https://www.roboticsproceedings.org/rss20/p107.html}
}

@inproceedings{ji2024exbody2,
  author={Ji, Mazeyu and Peng, Xuanbin and Liu, Fangchen and Li, Jialong and Yang, Ge and Cheng, Xuxin and Wang, Xiaolong},
  title={{ExBody2}: Advanced Expressive Humanoid Whole-Body Control},
  booktitle={2026 IEEE International Conference on Robotics and Automation (ICRA)},
  pages={613--620},
  year={2026},
  doi={10.1109/ICRA57385.2026.11696204},
  url={https://doi.org/10.1109/ICRA57385.2026.11696204}
}

@inproceedings{he2024omnih2o,
  author={He, Tairan and Luo, Zhengyi and He, Xialin and Xiao, Wenli and Zhang, Chong and Zhang, Weinan and Kitani, Kris M. and Liu, Changliu and Shi, Guanya},
  title={{OmniH2O}: Universal and Dexterous Human-to-Humanoid Whole-Body Teleoperation and Learning},
  booktitle={Proceedings of the 8th Conference on Robot Learning},
  series={Proceedings of Machine Learning Research},
  volume={270},
  pages={1516--1540},
  year={2025},
  publisher={PMLR},
  url={https://proceedings.mlr.press/v270/he25b.html}
}

@inproceedings{ze2025twist,
  author={Ze, Yanjie and Chen, Zixuan and Ara{\'u}jo, Jo{\~a}o Pedro and Cao, Zi-ang and Peng, Xue Bin and Wu, Jiajun and Liu, C. Karen},
  title={{TWIST}: Teleoperated Whole-Body Imitation System},
  booktitle={Proceedings of the 9th Conference on Robot Learning},
  series={Proceedings of Machine Learning Research},
  volume={305},
  pages={2143--2154},
  year={2025},
  publisher={PMLR},
  url={https://proceedings.mlr.press/v305/ze25a.html}
}

@inproceedings{ze2025twist2,
  author={Ze, Yanjie and Zhao, Siheng and Wang, Weizhuo and Kanazawa, Angjoo and Duan, Rocky and Abbeel, Pieter and Shi, Guanya and Wu, Jiajun and Liu, C. Karen},
  title={{TWIST2}: Scalable, Portable, and Holistic Humanoid Data Collection System},
  booktitle={2026 IEEE International Conference on Robotics and Automation (ICRA)},
  pages={15900--15907},
  year={2026},
  doi={10.1109/ICRA57385.2026.11696896},
  url={https://doi.org/10.1109/ICRA57385.2026.11696896}
}

@inproceedings{he2025asap,
  author={He, Tairan and Gao, Jiawei and Xiao, Wenli and Zhang, Yuanhang and Wang, Zi and Wang, Jiashun and Luo, Zhengyi and He, Guanqi and Sobanbabu, Nikhil and Pan, Chaoyi and Yi, Zeji and Qu, Guannan and Kitani, Kris and Hodgins, Jessica K. and Fan, Linxi and Zhu, Yuke and Liu, Changliu and Shi, Guanya},
  title={{ASAP}: Aligning Simulation and Real-World Physics for Learning Agile Humanoid Whole-Body Skills},
  booktitle={Proceedings of Robotics: Science and Systems},
  year={2025},
  doi={10.15607/RSS.2025.XXI.066},
  url={https://roboticsproceedings.org/rss21/p066.html}
}

@article{tessler2024maskedmimic,
  author={Tessler, Chen and Guo, Yunrong and Nabati, Ofir and Chechik, Gal and Peng, Xue Bin},
  title={{MaskedMimic}: Unified Physics-Based Character Control Through Masked Motion Inpainting},
  journal={ACM Transactions on Graphics},
  volume={43},
  number={6},
  pages={209:1--209:21},
  year={2024},
  doi={10.1145/3687951},
  url={https://doi.org/10.1145/3687951}
}

@inproceedings{he2024hover,
  author={He, Tairan and Xiao, Wenli and Lin, Toru and Luo, Zhengyi and Xu, Zhenjia and Jiang, Zhenyu and Kautz, Jan and Liu, Changliu and Shi, Guanya and Wang, Xiaolong and Fan, Linxi Jim and Zhu, Yuke},
  title={{HOVER}: Versatile Neural Whole-Body Controller for Humanoid Robots},
  booktitle={2025 IEEE International Conference on Robotics and Automation (ICRA)},
  pages={9989--9996},
  year={2025},
  doi={10.1109/ICRA55743.2025.11128549},
  url={https://doi.org/10.1109/ICRA55743.2025.11128549}
}

@inproceedings{li2025amo,
  author={Li, Jialong and Cheng, Xuxin and Huang, Tianshu and Yang, Shiqi and Qiu, Ri-Zhao and Wang, Xiaolong},
  title={{AMO}: Adaptive Motion Optimization for Hyper-Dexterous Humanoid Whole-Body Control},
  booktitle={Proceedings of Robotics: Science and Systems},
  year={2025},
  doi={10.15607/RSS.2025.XXI.061},
  url={https://www.roboticsproceedings.org/rss21/p061.html}
}

@inproceedings{chen2025gmt,
  author={Chen, Zixuan and Ji, Mazeyu and Cheng, Xuxin and Peng, Xuanbin and Peng, Xue Bin and Wang, Xiaolong},
  title={{GMT}: General Motion Tracking for Humanoid Whole-Body Control},
  booktitle={IEEE/RSJ International Conference on Intelligent Robots and Systems (IROS)},
  year={2026},
  note={Accepted for publication},
  url={https://arxiv.org/abs/2506.14770}
}

@article{luo2026sonic,
  author={Luo, Zhengyi and Yuan, Ye and Wang, Tingwu and Li, Chenran and Casta{\~n}eda, Fernando and Chen, Sirui and Cao, Zi-Ang and Li, Jiefeng and Minor, David and Ben, Qingwei and Park, Jinhyung and Sami, David and Wang, Zi and Da, Xingye and Ding, Runyu and Hogg, Cyrus and Song, Lina and Lim, Edy and Jeong, Eugene and He, Tairan and Xue, Haoru and Xiao, Wenli and Yuen, Simon and Kautz, Jan and Chang, Yan and Iqbal, Umar and Fan, Linxi "Jim" and Zhu, Yuke},
  title={{SONIC}: Supersizing Motion Tracking for Natural Humanoid Whole-Body Control},
  journal={Science Robotics},
  volume={11},
  number={117},
  year={2026},
  doi={10.1126/scirobotics.aed4592},
  note={Art. no. eaed4592},
  url={https://doi.org/10.1126/scirobotics.aed4592}
}

@misc{li2026omnitrack,
  author={Li, Yuhan and Zhi, Peiyuan and Wang, Yunshen and Liu, Tengyu and Yan, Sixu and Liu, Wenyu and Wang, Xinggang and Jia, Baoxiong and Huang, Siyuan},
  title={{OmniTrack}: General Motion Tracking via Physics-Consistent Reference},
  year={2026},
  doi={10.48550/arXiv.2602.23832},
  howpublished={arXiv:2602.23832},
  url={https://arxiv.org/abs/2602.23832}
}

@misc{li2026pgmt,
  author={Li, Hongyi and Li, Peizhuo and Tao, Yucheng and Wang, Ze and Xu, Fangzhou and Chen, Jinyi and Yuan, Yanyan and Jia, Dapeng and Jin, Yongbin and Fan, Mingfeng and Sartoretti, Guillaume and Wang, Hongtao},
  title={{PGMT}: Perceptive General Motion Tracking for Humanoid Robots},
  year={2026},
  doi={10.48550/arXiv.2609.08511},
  howpublished={arXiv:2609.08511},
  url={https://arxiv.org/abs/2609.08511}
}

@misc{amadio2026tracking,
  author={Amadio, Fabio and Mingo Hoffman, Enrico},
  title={What Matters in Humanoid General Motion Tracking? {An} Empirical Study},
  year={2026},
  doi={10.48550/arXiv.2607.19903},
  howpublished={arXiv:2607.19903},
  url={https://arxiv.org/abs/2607.19903}
}

@inproceedings{guo2024momask,
  author={Guo, Chuan and Mu, Yuxuan and Javed, Muhammad Gohar and Wang, Sen and Cheng, Li},
  title={{MoMask}: Generative Masked Modeling of {3D} Human Motions},
  booktitle={Proceedings of the IEEE/CVF Conference on Computer Vision and Pattern Recognition},
  pages={1900--1910},
  year={2024},
  doi={10.1109/CVPR52733.2024.00186},
  url={https://doi.org/10.1109/CVPR52733.2024.00186}
}

@inproceedings{cohan2024condmdi,
  author={Cohan, Setareh and Tevet, Guy and Reda, Daniele and Peng, Xue Bin and van de Panne, Michiel},
  title={Flexible Motion In-betweening with Diffusion Models},
  booktitle={ACM SIGGRAPH 2024 Conference Papers},
  pages={1--9},
  year={2024},
  doi={10.1145/3641519.3657414},
  url={https://doi.org/10.1145/3641519.3657414}
}

@inproceedings{zhao2025dart,
  author={Zhao, Kaifeng and Li, Gen and Tang, Siyu},
  title={{DartControl}: A Diffusion-Based Autoregressive Motion Model for Real-Time Text-Driven Motion Control},
  booktitle={International Conference on Learning Representations},
  year={2025},
  url={https://proceedings.iclr.cc/paper_files/paper/2025/hash/3b057de5a2e38bd8fa10201866c20dbf-Abstract-Conference.html}
}

@inproceedings{tevet2025closd,
  author={Tevet, Guy and Raab, Sigal and Cohan, Setareh and Reda, Daniele and Luo, Zhengyi and Peng, Xue Bin and Bermano, Amit H. and van de Panne, Michiel},
  title={{CLoSD}: Closing the Loop between Simulation and Diffusion for multi-task character control},
  booktitle={International Conference on Learning Representations},
  year={2025},
  url={https://proceedings.iclr.cc/paper_files/paper/2025/hash/7319b7561ffe5e2f6419acd4a2f52d6b-Abstract-Conference.html}
}

@inproceedings{wang2026fast,
  author={Zepeng Wang and Jiangxing Wang and Shiqing Yao and Yu Zhang and Ziluo Ding and Ming Yang and Yuxuan Wang and Haobin Jiang and Chao Ma and Xiaochuan Shi and Zongqing Lu},
  title={General Humanoid Whole-Body Control via Pretraining and Fast Adaptation},
  booktitle={Conference on Robot Learning (CoRL)},
  year={2026},
  note={Accepted for publication},
  url={https://openreview.net/forum?id=HWaigsXLaH}
}

@IEEEtranBSTCTL{IEEEreferenceStyle,
  CTLuse_forced_etal={no}
}
\end{document}